\documentclass[10pt, conference]{ieeeconf}
\IEEEoverridecommandlockouts
\usepackage{cite}
\usepackage{amsmath,amssymb,amsfonts}
\usepackage{mathtools}
\usepackage{graphicx}
\usepackage{textcomp}
\usepackage{xcolor}
\usepackage[ruled,vlined, linesnumbered]{algorithm2e}
\usepackage{stackengine}  
\usepackage{booktabs}
\usepackage[skip=10pt, font=footnotesize]{caption}
\usepackage{siunitx}
\usepackage{hyperref}  
\hypersetup{
    colorlinks,
    linkcolor={red!75!black},
    citecolor={blue!75!black},
    urlcolor={blue!80!black}
}

\usepackage{color}
\usepackage{xcolor}

\usepackage{cancel}

\newcommand{\X}{
\mathbf{X}
}

\newcommand{\y}{
\mathbf{y}
}

\newcommand{\f}{
\mathbf{f}
}

\newcommand{\K}{
\mathbf{K}
}

\newcommand{\li} {
{(i)}
}

\newcommand{\Q}{
\mathbf{Q}
}

\newcommand{\Lm}{
\mathbf{\Lambda}
}

\newcommand{\I}{
\mathbf{I}
}

\newcommand{\Rn}{
\mathbb{R}
}

\newcommand{\R}{
\mathbf{R}
}

\newcommand{\bigOgpmem} {
\mathcal{O}(n^2)
}
\newcommand{\bigOgp} {
\mathcal{O}(n^3)
}
\newcommand{\bigOgpmean} {
\mathcal{O}(n)
}

\newcommand{\bigOsgpmem} {
\mathcal{O}(nm)
}
\newcommand{\bigOsgp} {
\mathcal{O}(nm^2)
}
\newcommand{\bigOsgpmean} {
\mathcal{O}(m)
}
\newcommand{\bigOsgpvar} {
\mathcal{O}(m^2)
}

\newcommand{\bigOlgpmem} {
\mathcal{O}(Mp^2)
}

\newcommand{\bigOlgppred} {
\mathcal{O}(Np^3)
}
\newcommand{\bigOlgpmean} {
\mathcal{O}(Np)
}

\newcommand{\bigOmsgpmem} {
\mathcal{O}(Mpu)
}
\newcommand{\bigOmsgp} {
\mathcal{O}(Mpu^2)
}
\newcommand{\bigOmsgppred} {
\mathcal{O}(Npu^2)
}
\newcommand{\bigOmsgpmean} {
\mathcal{O}(Nu)
}

\begin{document}

\title{Multi-Sparse Gaussian Process: Learning based Semi-Parametric Control}

\author{Mouhyemen Khan, Akash Patel$^{\dagger}$, and Abhijit Chatterjee
\thanks{Mouhyemen Khan and Abhijit Chatterjee are with the School of Electrical and Computer Engineering, $^{\dagger}$Akash Patel is with the Institute of Robotics and Intelligent Machines, Georgia Institute of Technology, Atlanta, GA 30332, USA: {\tt \{mouhyemen.khan, abhijit.chatterjee, apatel435\}@gatech.edu }
}
}

\maketitle

\begin{abstract}  
A key challenge with controlling complex dynamical systems is to accurately model them.
However, this requirement is very hard to satisfy in practice.
Data-driven approaches such as Gaussian processes (GPs) have proved quite effective by employing regression based methods to capture the unmodeled dynamical effects.
However, GPs scale cubically with data, and is often a challenge to perform real-time regression.
In this paper, we propose a semi-parametric framework exploiting sparsity for learning-based control. We combine the parametric model of the system with multiple sparse GP models to capture any unmodeled dynamics.
\textit{Multi-Sparse Gaussian Process (MSGP)} divides the original dataset into multiple sparse models with unique hyperparameters for each model. Thereby, preserving the richness and uniqueness of each sparse model. For a query point, a weighted sparse posterior prediction is performed based on $N$ neighboring sparse models. Hence, the prediction complexity is significantly reduced from $\bigOgp$ to $\bigOmsgppred$, where $p$ and $u$ are data points and pseudo-inputs respectively for each sparse model.
We validate MSGP's learning performance for a quadrotor using a geometric controller in simulation. 
Comparison with GP, sparse GP, and local GP shows that MSGP has higher prediction accuracy than sparse and local GP, while significantly lower time complexity than all three. We also validate MSGP on a hardware quadrotor for unmodeled mass, inertia, and disturbances. The experiment video can be seen at: \url{https://youtu.be/zUk1ISux6ao}

\end{abstract}

\section{INTRODUCTION}\label{sec:intro}
Precise knowledge of the model for controlling complex nonlinear systems is crucial for achieving state estimation \cite{Ko2009_state_estimation} and robot tracking control \cite{Tuong2008_torque_control}. However, it is often difficult, if not impossible, to accurately model such systems with high fidelity. Data-driven based methods have been successful in learning the unmodeled components of system dynamics to a high degree in a supervised learning paradigm. One popular approach is using Gaussian processes (GPs) for non-parametric regression \cite{Rasmussen2003_gpml, Tuong2010_gp_control}. Performing real-time regression with GPs is challenging due to its complexity of $\bigOgp$, since matrix inverse operations are performed on $n$ observed data points. This limits use of GPs in applications requiring large amounts of data or fast computation times, e.g. safety-critical autonomous vehicles and aerial drones. Approximations to GPs, both sparse and local, have been proposed to deal with these challenges while balancing the inherent trade-off between accuracy and complexity. In this paper, we propose a novel framework that leverages benefits from both sparse and local GP approximations to perform low-complexity, more efficient, and accurate learning in robot control called \textit{Multi-Sparse Gaussian Process} (MSGP).

Various non-parametric regression frameworks for real-time learning have been proposed. Locally weighted projection regression (LWPR) learns the true function locally, spanned by a number of univariate regressions with weighted kernels \cite{Vijayakumar2000_lwpr}. LWPR, however, requires manual tuning of many metaparameters and a large number of linear models for achieving satisfactory approximation of the original function. GPs offer an appealing alternative to LWPR for model learning. GPs are flexible since they learn the model structure and estimate any hyperparameters from the data itself \cite{Rasmussen2003_gpml}. Thus, GPs are very powerful in capturing higher order nonlinearities with high prediction accuracy, e.g., in robot tracking and control \cite{Tuong2010_gp_control}.

Due to GP's limitation on large datasets owing to its time complexity, many extensions have been developed. Local GP (LGP) is a hybrid between GP and LWPR  that divides the entire dataset into local models and predicts using weighted average of these local models \cite{Tuong2009_local_gp}. LGP is shown to outperform LWPR while retaining accuracy close to standard GP. However, LGP suffers from not retaining the uniqueness of each model and as the size per model increases, so does the complexity since it assumes a full GP per local model. Unlike LGP, sparse techniques approximate GPs by selecting a set of pseudo-inputs to mimic the original likelihood. There are many sparse approximations and we refer the reader to \cite{Bui2017_unifying, Quinonero2005_unifying} regarding details and their unifying framework. An online mixture of experts using sparse GPs has been used for learning shared control policies \cite{Soh_sparse_gp}. This study applies their learning based control on a smart wheelchair which is not dynamically safety critical. Learning-based control using GPs has been demonstrated for quadrotors in \cite{Felix_GPquad}, \cite{Wang_GPquad}, \cite{Sheran_GPquad}, \cite{Cao_GPquad}, \cite{Smith_GPquad}. However, they use standard GPs in their framework and are not scalable to large datasets.

Our key \textbf{contributions} are the following.
First, we present a semi-parametric framework using sparsity that estimates model nonlinearities. To this end, multiple sparse GPs are equipped with basis functions obtained from physics first principles. Semi-parametric methods have been applied for inverse dynamics \cite{Tuong2010_gp_control}, \cite{Romeres2016_semiparametric_inverse}, system identification \cite{Wu2012_semiparametric_systemid}, and forward dynamics \cite{Romeres2016_semiparametric_forward}; all using standard GPs. To the best of our knowledge, no prior work has merged semi-parametric modeling using sparse approximations of GPs.
Second, we create multiple sparse approximations of the original GP clustered into regionally sparse models without making any global assumptions. Local models hinder prediction accuracy at the benefit of reduced complexity. To overcome this, each sparse model is optimized for its own hyperparameters and a weighted sparse posterior prediction is performed.
Third, we validate the learning performance of MSGP on a hardware quadrotor platform. Learning-based control \textit{especially using sparsity for a safety-critical system such as a quadrotor} has not been demonstrated before to the best of our knowledge. We address sparse based learning on a quadrotor, whose dynamics evolve in the tangent bundle to $SE(3)$. 
 Additionally, we also compare the learning performance of MSGP against other GP methods on a geometric quadrotor controller \cite{Lee2010_geometric} in simulation.

The rest of the paper is organized as follows. Section \ref{sec:problem} presents the problem formulation. Section \ref{sec:prelim} covers background regarding GPs and sparse GPs. Section \ref{sec:msgp} describes the proposed MSGP framework. Semi-parametric based control with MSGP is shown in Section \ref{sec:quadrotor}. Simulation results showing MSGP's learning performance are discussed in Section \ref{sec:simulation}. Hardware experiments using MSGP is discussed in Section \ref{sec:hardware} followed by the conclusion in Section \ref{sec:conclusion}.

\section{PROBLEM STATEMENT}\label{sec:problem}
We consider a nonlinear, continuous-time system,
\begin{align}\label{eq:nonlinear_dynamics}
\dot{x} &= \underbrace{f(x(t), u(t))}_{\text{parametric}} + \underbrace{g( x(t), u(t) )}_{\text{non-parametric}},
\end{align}
where $x(t) \in \mathbb{R}^n$ is the state and $u(t) \in \mathbb{R}^m$ is the control input at time $t$. The system dynamics is divided into a known parametric model $f(x,u)$ and an unknown non-parametric model $g(x, u)$. The latter contains the unmodeled dynamics.

The goal is to estimate the model nonlinearities for (\ref{eq:nonlinear_dynamics}) in a \textit{semi-parametric manner using sparsity} by placing multiple sparse GP priors on the non-parametric component. The basis functions for these sparse GPs take into account the physical knowledge of the system, i.e., the parametric component. This is equivalent to a semi-parametric model given by,
\begin{align}
\dot{x} &\sim f(x, u) + \Sigma_i^M \mathcal{SGP} \ \big( \mathbf{0} , \ k(x,x') \ \big), \label{eq:sum_of_sparse}\\
		&\sim \mathcal{MSGP} \ \big( \ f(x, u) \ , \ k(x,x') \ \big) \label{eq:semiparam_msgp},
\end{align}
Comparing the resulting dynamics in (\ref{eq:semiparam_msgp}) with (\ref{eq:nonlinear_dynamics}) and (\ref{eq:sum_of_sparse}) makes it clear that \textit{the objective of MSGP is to model the nonlinearities using multiple sparse GPs}. This problem has been addressed before using standard GPs. We differ in our motivation to achieve the same using multiple sparse approximations to GPs, without comprising speed and accuracy for growing datasets.
We assume that we can measure, $\hat{g}(x, u) = \dot{x} - f(x,u) + \epsilon$, which are corrupted by independent, zero-mean, and bounded noise, $||\epsilon|| \leq \sigma$. 
We also assume a nominal controller $u_{nom}(t)$ exists that drives the parametric model $f(\cdot)$ to the zero equilibrium point.

\section{BACKGROUND PRELIMINARIES}\label{sec:prelim}
Here, we present the preliminaries for GPs and one of its sparse variants called sparse pseudo-input GP (SPGP) \cite{Snelson2006_spgp}.

\subsection{Standard GP Regression}\label{subsec:gpr}
GPs are a popular choice for nonparametric regression in machine learning. We are interested in learning an underlying latent function $g(x)$, for which we assume to have noisy observations given by, $y_i = g(x_i) + \epsilon_i$, where $\epsilon_i \sim \mathcal{N}(0, \sigma_{\omega}^2)$. Given a set of $n$ training data points, with input vectors $x \in \mathbb{R}^d$, and scalar noisy observations $y \in \mathbb{R}$, we compose the dataset:  $\mathcal{D}_n = \{ \X_n, \y_n \}$, where $\X_n = \{x_i\}_{i=1}^{n}$ and $\y_n = \{y_i\}_{i=1}^n$. A GP places a distribution on the unknown function $g(x)$, treating it as random variables associated with different values of $x$, any finite number of which produces a consistent joint Gaussian distribution \cite{Rasmussen2003_gpml}. For instance, $x$ here could represent a robot's states, and $g(x)$ could represent the unmodeled system dynamics (see Section \ref{subsec:msgp_controller}).

A GP can be fully specified by its mean $\mu(x)$ and covariance $k(x,x')$. The latter is also called the kernel, measuring similarity between any two inputs $x, x'$. GPs can be used to predict the function value, $g(x_*)$, for an arbitrary query point $x_*$, by conditioning on previous observations. The posterior predictive mean and variance are then given by \cite{Rasmussen2003_gpml}:
\begin{align}\label{eq:gp}
\mu(x_*) &= k^{\top}_{n*} \big( \K_n + \sigma_{\omega}^2 \I_n \big)^{-1} \y_n \\
\sigma(x_*)^2 &= k (x_*, x_*) - k^{\top}_{n*} \big( \K_n + \sigma_{\omega}^2 \I_n \big)^{-1} k_{n*},
\end{align}
where $k_{n*} = \big[k(x_1, x_*), \ldots , k(x_n, x_*) \big]^{\top}$ is the covariance between the input points in $\X_n$ and query point $x_*$, $\K_n \in \mathbb{R}^{n \times n}$ has entries $\big[ \K_n \big]_{(i,j)} = k(x_i, x_j), \ i, j \in \{1, \ldots, n\}$, is the covariance matrix between pairs of input points, and $\I_n \in \mathbb{R}^{n \times n}$ is the identity matrix. The hyperparameters of a GP depend on the kernel choice and can be problem-dependent. We refer the reader to \cite{Rasmussen2003_gpml} for a review of different kernels. For a Gaussian kernel, the hyperparameters that best suit the particular dataset can be derived by maximizing the log marginal likelihood using quasi-Newton methods  \cite{Rasmussen2003_gpml}.

\subsection{Sparse Pseudo-Input GP Regression}\label{subsec:sgpr}
Despite GPs being very powerful regressors, as the dataset grows larger, they become computationally intractable. 
Hence, many sparse approximations of GPs have been developed to bring down the complexity cost while retaining accuracy \cite{Snelson2006_spgp}, \cite{Csato2002_online_sgp}, \cite{Seeger2003_ivm}, \cite{Minka2001_pep}, \cite{Titsias2009_vfe}. Broadly stated, sparse approximations of GPs fall in two major categories: \textit{approximate generative model with exact inference} and \textit{exact generative model with approximate inference}. Unifying theories for these various frameworks are discussed in \cite{Bui2017_unifying}, \cite{Quinonero2005_unifying}. We focus on a variant of the former category: \textit{Sparse Pseudo-Input Gaussian Process} (SPGP) \cite{Snelson2006_spgp}. 

The starting point to any GP approximation method is through a set of so-called \textit{inducing} or \textit{pseudo points} giving rise to sparsity. Consider a pseudo-dataset $\mathcal{D}_m = \big[ \bar{\X}_m, \bar{\y}_m \big]$, $m \ll n$: pseudo-inputs are $\bar{\X}_m = \{ \bar{x} \}_{i=1}^m$ and pseudo-targets are $\bar{\y}_m = \{ y_i \}_{i=1}^m$. The objective is to find the posterior distribution over pseudo targets, followed by the prediction distribution by integrating the likelihood with the posterior. 
A complete mathematical treatment for the derivation of the predictive distribution of SPGP can be found in \cite{Snelson2006_spgp}. Here, we simply present predictive mean and variance of SPGP:
\begin{align}\label{eq:sgp}
\mu(x_*) &= k^{\top}_{m*} \Q_m^{-1} \big( \Lm_n + \sigma_{\omega}^2 \I_n \big)^{-1} \y_n \\
\sigma(x_*)^2 &= k (x_*, x_*) - k^{\top}_{m*} \big( \K_m^{-1} - \Q_m^{-1} \big) k_{m*} + \sigma_{\omega}^2,
\end{align}
where $\big[ \K_{nm} \big]_{(i,j)} = k(x_i, \bar{x}_j), \ i \in \{1, ... , n \}, j \in \{1, ... , m\}$ is the covariance matrix between input points $\X_n$ and pseudo-inputs $\bar{\X}_m$, $\K_m \in \mathbb{R}^{m \times m}$ is the covariance between pairs of pseudo-inputs, $\Lm_n = \text{diag} \big[ \K_n \ - \  \K_{nm} \K_m^{-1} \K_{mn} \big]$ is a diagonal matrix, and $\Q_m = \K_m + \K_{nm}^{\top} \big( \Lm_n + \sigma_{\omega}^2 \I_n  \big)^{-1} \K_{nm}$. Inversion cost for covariance matrix is reduced to $\bigOsgp$ \cite{Snelson2006_spgp}. The cost per test case is $\bigOsgpmean$ and $\bigOsgpvar$ for predictive mean and variance respectively.

\section{MULTI-SPARSE GAUSSIAN PROCESS}\label{sec:msgp}
We discuss our proposed methodology inspired from the theoretical developments of SPGP and architecture of LGP. At the onset, MSGP can be seen simply as the combination of SPGP and LGP, however, it outperforms both SPGP and LGP which is very counter intuitive. MSGP at its core is different from LGP by using multiple sparse models (instead of full GP models) and unique hyperparameters in each model. Note that, although we choose SPGP as the sparse GP representative, MSGP is agnostic to the underlying sparse approximation.


\subsection{Multi-Sparse Model Clustering}\label{subsec:data_loc}
\begin{figure}[!b]
\vspace{-.35cm}
\centering
\includegraphics[width=0.65\linewidth]{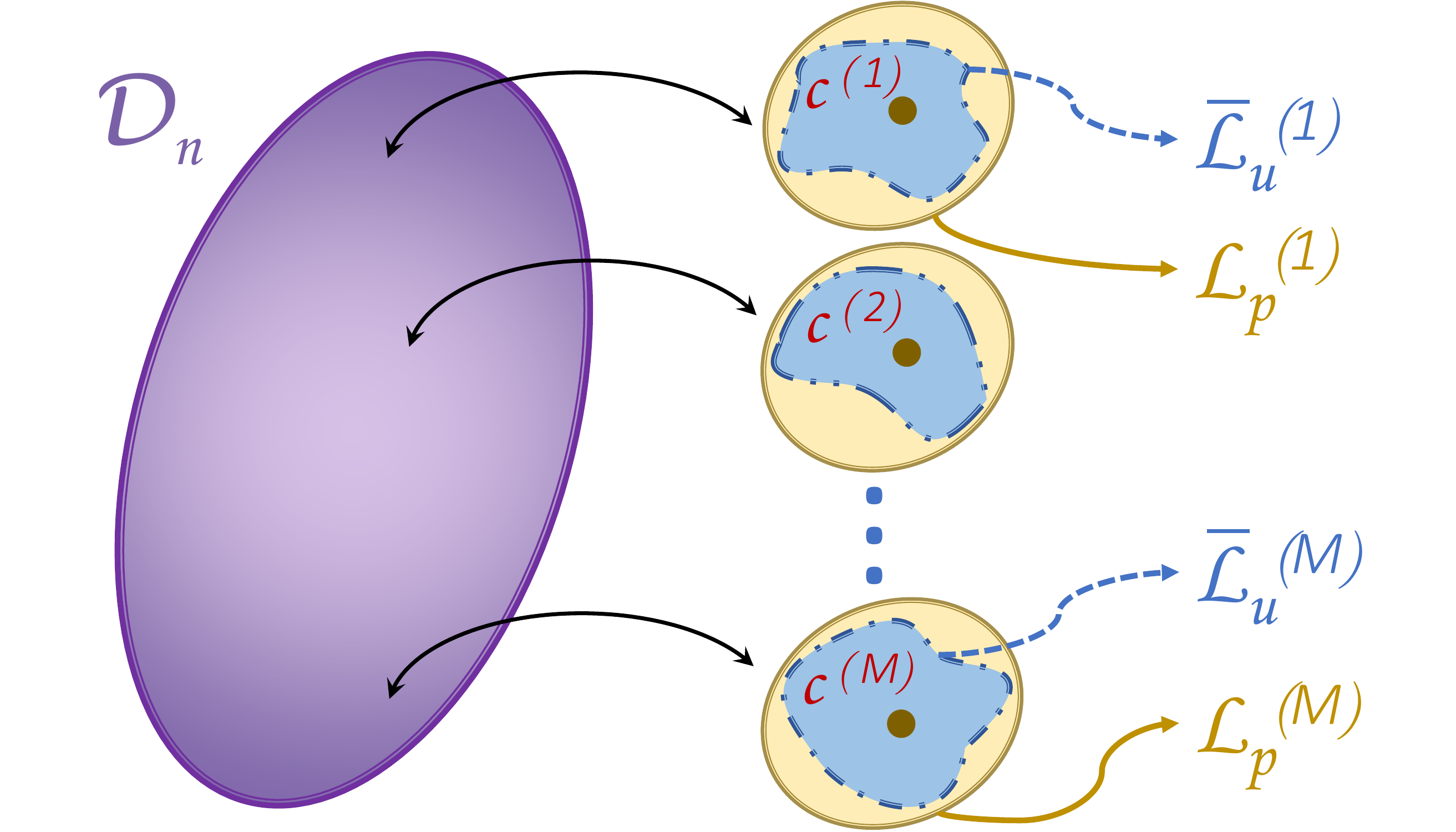}
\caption{The original dataset $\mathcal{D}_n$ (purple) is divided into $M$ local models $\mathcal{L}_p$ (yellow) with approximately $p$ data points each and a corresponding center $c$. Each local model is further approximated into its sparse representation $\bar{\mathcal{L}}_u$ (blue), with $u \ll p$ local pseudo-inputs.}
\label{fig:sparse_model_clustering}
\end{figure}

The entire dataset of $n$ training points is divided into $M$ local models (randomly/deterministically), $\mathcal{L}_p$, each with approximately $p$ data points. 
Every $i^{th}$ local model is denoted by $\mathcal{L}_p^\li$, comprising of its dataset, $\{\X_p, \y_p \}^\li$  and corresponding center, $c^\li = \text{mean}(\X_p^\li)$. 
Next, sparsity is introduced in each local model by selecting a set of pseudo-inputs $\bar{\X}_u^\li \in \X_p^\li$, where $u \ll p$. These pseudo-inputs are selected arbitrarily at first for each sparse model; to be optimized later in hyperparameter tuning phase. Thus, each sparse model in MSGP (see Figure \ref{fig:sparse_model_clustering}) is specified as $\bar{\mathcal{L}}_u^\li = \{ \X_p, \y_p, c, \bar{\X}_u \}^\li$; parameterized by localized hyperparameters $\bar{\Theta}^\li = [\sigma_{\omega}^2, \sigma_{f}^2, l ]^\li$, where $\sigma_{\omega}$ and $\sigma_{f}$ are noise and signal variance, and $l$ is characteristic length-scale.

\subsection{Localized Hyperparameter Tuning}\label{subsec:hyp_loc}
Each sparse model is parameterized by $\bar{\Theta}^\li$ that best fits its own dataset since we perform an optimization procedure on each model. The marginal likelihood for each model is,
\begin{align}\label{eq:local_marg_llh}
p \big( \y_p | \X_p, \bar{\X}_u, \bar{\Theta} \big) &= \int p \big( \y_p \ | \ \X_p, \bar{\X}_u, \bar{\f} \big) p \big( \bar{\f} \ | \ \bar{\X}_u \big) d \bar{\f} \notag \\
p (\cdot ) &= \mathcal{N} \big( \y_p \ | \ \mathbf{0} \ , \ \mathbf{L}_p + \Lm_p + \sigma_{\omega}^2 \I_p \big) \notag \\
&= \mathcal{N} \big( \y_p \ | \ \mathbf{0} \ , \ \bar{\K}_p \big),
\end{align}
where $\mathbf{L}_p := \K_{pu} \K_{u}^{-1} \K_{up}$, $\big[ \K_{pu} \big]_{(i,j)} = k(x_i, \bar{x}_j), i \in \{1, ..., p \}, j \in \{1, ..., u\}$ is the covariance between local inputs $\X_p$ and pseudo-inputs $\bar{\X}_u$, $\K_u \in \mathbb{R}^{u \times u}$ is the covariance between pairs of local pseudo-inputs. $\Lm_p = \text{diag} \big[ \K_p \ - \mathbf{L}_p\big]$ is a diagonal matrix, and $\bar{\K}_p := \mathbf{L}_p + \Lm_p + \sigma_{\omega}^2 \I_p$.

By maximizing the log marginal likelihood of (\ref{eq:local_marg_llh}), we can jointly optimize for the hyperparameters, $\bar{\Theta}^\li$, and pseudo-inputs, $\bar{\X}_u^\li$, for each sparse model as given by:
\begin{align}\label{eq:local_log_marg_llh}
\log p (\cdot ) &= 
- \frac{1}{2}\Big(p \log(2\pi) - \log | \bar{\K}_p | - \y_p \bar{\K}_p ^ {-1} \y_p \Big). 
\end{align}
The approximate multiple sparse generative model has attractive properties. Firstly, the training complexity in MSGP has been reduced to $\bigOmsgp$ from GP's $\bigOgp$, which is a significant reduction.
Moreover, by optimizing the hyperparameters along with the pseudo-inputs for each sparse model, we preserve the richness and uniqueness of each model unlike LGP.
Next, we look at sparse posterior predictions for a query point $x^*$.

\subsection{Multi-Sparse Posterior Prediction}\label{subsec:pred_loc}
Optimizing hyperparameters for each sparse model may give rise to overfitting during the prediction phase. To remedy this, the posterior prediction in MSGP uses a weighted averaging over $N$ neighboring sparse predictions $\hat{\mu}$ for a query point $x_*$. The idea of weighted averaging for predictors was first introduced in LWPR; also used by LGP for its predictions.
Akin to LWPR, we also perform weighted averaging, but using weighted sparse posterior predictions instead. The $N$ nearest sparse models can be determined quickly using the Gaussian kernel:
\begin{align}
d(x_*, c)^\li &= \text{exp} \Big( - \frac{ || (x_* - c^\li) || ^2 }{ 2 \ l^\li \cdot l^\li } \Big),
\end{align}
where $c^\li$ is the center of model $\mathcal{L}_p^\li$ and $l^\li$ is the respective characteristic length scale from model $\mathcal{L}_p^\li$. Finally, the posterior prediction of MSGP's predictive mean is as follows:
\begin{align}
\hat{\mu}(x_*) &= \frac{\sum_{j=1}^{N} d_j^\li \mu(x_*)_j}{ \sum_{j=1}^{N} d_j}, \label{eq:weighted_mean} \\
\mu(x_*)_j &= k^{\top}_{u*} \Q_u^{-1} \big( \Lm_p + {}^\li \sigma_{\omega}^2 \I_p \big)^{-1} \y_p \label{eq:msgp_mean} , 
\end{align}
where $k_{u*} = \big[k(x_1, x_*), \ldots , k(x_u, x_*) \big]^{\top}$ gives the covariance between the local pseudo-inputs in $\bar{\X}_p$ and query point $x_*$, and ${}^\li \sigma_{\omega}^2$ is the local noise variance. Hence, the predictive mean complexity in MSGP is $\bigOmsgppred$ compared to LGP's complexity of $\bigOlgppred$.

\section{QUADROTOR LEARNING AND CONTROL}\label{sec:quadrotor}
Most modern controllers require accurate knowledge of the model for improved trajectory tracking. By learning the unmodeled component using MSGP, we demonstrate improved trajectory tracking for a quadrotor in $SE(3)$. First, we briefly outline the dynamics and the geometric controller of the quadrotor. We then discuss the augmentation of MSGP with the geometric controller for improved tracking in presence of unmodeled dynamics and uncertainties.


\subsection{Geometric Dynamics Model}\label{subsec:dynamics}
We consider the complete dynamics of a quadrotor model evolving in a coordinate-free framework. This framework uses a geometric representation for its attitude given by a rotation matrix $\R$ in $SO(3) := \{ \R \in \Rn^{3 \times 3} \ | \ \R^{\top}\R = \I, \ \det(\R) = 1 \}$. $\R$ represents the rotation from body-frame to the inertial-frame. The origin of the body-frame is given by the quadrotor's center of mass, denoted by $r \in \Rn^3$. 
A quadrotor is underactuated since it has $6$ DOF, due to its configuration space $\mathcal{C} := SE(3)$, but $4$ control inputs; thrust $F \in \Rn$ and moment $M \in \Rn^3$. The equations of motion are:
\begin{align}
\dot{r} 		&= v 	\label{eq:dyn1} \\
m\dot{v}		&= m g e_3 - F \R e_3 	\label{eq:dyn2}\\
\dot{\R}		&= \R \Omega^{\times}	\label{eq:dyn3}\\
J\dot{\Omega}	&= M - (\Omega \times J \Omega)	\label{eq:dyn4}
\end{align}
where $v$ is velocity in inertial frame, $(\cdot)^{\times}: \Rn^3 \rightarrow so(3)$ denotes the skew-symmetric operator, $\forall x,y \in \Rn^3, x^{\times}y = x \times y$ , $m$ is mass, $g$ is gravity, $J \in \Rn^{3 \times 3}$ is inertia matrix, $e_3 = [0 \ 0 \ 1]^{\top}$, and $\Omega \in \Rn^3$ is body-frame angular velocity.

\subsection{Geometric Controller Tracking in SE(3)}\label{subsec:controller}
The geometric controller used for trajectory tracking presented in \cite{Lee2010_geometric} has almost global exponential stability. This implies the quadrotor can reach any desired state in the state-space from any initial configuration. For a complete mathematical treatment for the nominal controller, see \cite{Lee2010_geometric}. Here, we just present the equations for nominal $F$ and $M$:
\begin{align}\label{eq:nom_controller}
\begin{rcases}
F &= (-k_r e_r - k_v e_v + mge_3 + m\ddot{r}_d)^{\top} \R e_3 	\\
M &= - k_\R e_\R - k_{\Omega} e_{\Omega} + \Omega \times J\Omega 				\\ 
& \ \ \ - J(\Omega^{\times} \R^{\top}\R_d \Omega_d - \R^{\top}\R_d \dot{\Omega}_d)  ,
\end{rcases}
\end{align}
where $k_{(\cdot)}$ are positive constants, $e_r = r - r_d$, $e_v = v - \dot{r}_d$, $e_{\R}		= \frac{1}{2}(\R_d^{\top}\R - \R^{\top}\R_d)^\vee$, and $e_{\Omega}	= \Omega - \R^{\top}\R_d\Omega_d$. The desired position, velocity, attitude, and angular acceleration are $r_d, \dot{r}_d, \R_d,$ and $\dot{\Omega}_d$ respectively. $(\cdot)^\vee$ is the inverse of  $(\cdot)^{\times}$, i.e. $(a^{\times})^\vee = a$.

\subsection{Learning based Control using MSGP}\label{subsec:msgp_controller}
The dynamical model in (\ref{eq:dyn1} - \ref{eq:dyn4}) and the controller presented in (\ref{eq:nom_controller}) deal with a precise model of the quadrotor. However, it is often difficult to accurately parameterize a dynamical system using physics first principles. Moreover, the model (\ref{eq:dyn1} - \ref{eq:dyn4}) does not consider aerodynamic drag, damping, wind effects, or time-varying changes to mass and inertia. Here, we will use MSGP to capture and learn any unmodeled effects to the system. Since unmodeled nonlinearities appear in the dynamics (\ref{eq:dyn2},\ref{eq:dyn4}). We use a total of six MSGPs, placing a prior on each dimension of the unmodeled state-space as shown below:
\begin{align}
m\dot{\hat{v}}			&=  m g e_3 - F \R e_3 	 \ + 	
\begin{bmatrix}
\ \mathcal{MSGP}_1(0, k(q,q'))  \\
\ \mathcal{MSGP}_2(0, k(q,q'))  \\
\ \mathcal{MSGP}_3(0, k(q,q')) 
\end{bmatrix} 								\label{eq:dyn2_msgp}\\
J \dot{\hat{\Omega}}	&=  M - (\hat{\Omega} \times J \hat{\Omega} )	+ 
\begin{bmatrix}
\ \mathcal{MSGP}_4(0, k(q,q'))  \\
\ \mathcal{MSGP}_5(0, k(q,q'))  \\
\ \mathcal{MSGP}_6(0, k(q,q')) 
\end{bmatrix}
. 								\label{eq:dyn4_msgp}
\end{align}
The input to MSGPs are $q = [r^{\top}, \dot{r}^{\top}, \Omega^{\top}]^{\top}$ and the target observations are given by the difference between (\ref{eq:dyn2}, \ref{eq:dyn4}) and (\ref{eq:dyn2_msgp}, \ref{eq:dyn4_msgp}), $\hat{y} = [m ( \dot{\hat{v}}- \dot{v} ) ^{\top}, J ( \dot{\hat{\Omega}} - \dot{\Omega})^{\top} ]^{\top} + \epsilon$. Given the input samples and noisy observations, this constitutes a proper regression problem. After learning, we can therefore perform prediction at a new query point $q_*$, where the sparse predictive mean of the unmodeled dynamics is calculated using (\ref{eq:weighted_mean}). This predictive mean is then used to modify the controller (\ref{eq:nom_controller}) with the learned dynamics as shown below,
\begin{align}\label{eq:msgp_controller}
\begin{rcases}
	F_{\text{MSGP}} &= F - m [\mu_1(q_*), \ \mu_2(q_*), \ \mu_3(q_*) ] \R e_3 	\\
	M_{\text{MSGP}} &= M - J[\mu_4(q_*), \ \mu_5(q_*), \ \mu_6(q_*) ]^{\top}.
\end{rcases}
\end{align}

\section{SIMULATION RESULTS}\label{sec:simulation}
We validate the MSGP semi-parametric learning framework by modifying the nominal controller's feedforward component on several test cases. We compare the MSGP's learning performance against the nominal, standard GP, SPGP, and LGP based controllers empirically. The \texttt{GPML} library in MATLAB is used for hyperparameter tuning and covariance calculations \cite{gpml_toolbox}. 

Desired trajectories are sinusoids where position reference is $[x_d, \ y_d, \ z_d]^{\top} = [4 \sin(0.8 t), 5 \sin(0.4 t), 2 \sin(0.4 t)]^{\top}$ and desired yaw is $\psi_d(t) = \text{atan2}(y_d,x_d)$, for $t \in [t_0, t_f]$. Nominal parameters are $m = 1.25$ $\si\kilogram$, $J = \text{diag}[1.1,  \ 1.1, \ 2.2]$ $\si\kilogram \si\meter^2$, gains $k_r = \text{diag}[5, 5, 5]$, $k_v = \text{diag}[0.5, 0.5, 2.0]$, $k_{\Omega} = \text{diag}[5, 10, 20]$, $k_{\R} = \text{diag}[30, 30, 30]$. In simulation, the quadrotor is subjected to these trajectories under model uncertainties using a low-gain nominal controller (\ref{eq:nom_controller}). As a result, the learned controller has a stronger effect to compensate for unmodeled dynamics by adjusting controller's feedforward component.

The unmodeled dynamics are broadly classified into three categories for investigation: \textit{parametric}, \textit{non-parametric}, and combined \textit{parametric and non-parametric}. For each category, we collect over $15500$ samples for training and over $6000$ samples for testing. 
One-tenth of the samples are chosen as pseudo-inputs for SPGP. Localization of data samples, into local models $\mathcal{L}_p$, for LGP and MSGP are done in the same state space as inputs to the respective GPs, i.e. $q = [r^{\top}, \dot{r}^{\top}, \Omega^{\top}]^{\top}$. Each local model consists of a maximum of $750$ samples resulting in over $20$ regional models. MSGP is further sparsed using one-fifth of the local samples as local pseudo-inputs.

\subsection{Parametric Unmodeled Dynamics}\label{subsec:param_error} 
\begin{figure*}[!t]
\centering
\includegraphics[width=1\linewidth, height=0.25\linewidth]{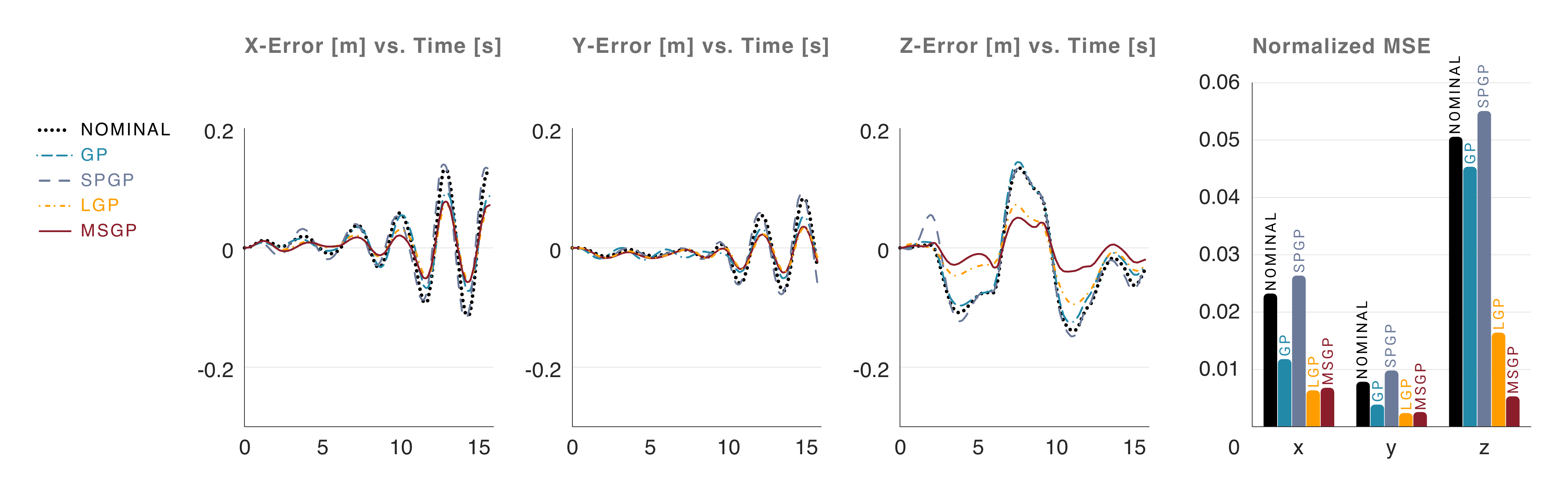}
\vspace{-0.8cm}
\caption{\textit{Parametric effects}: Tracking error between nominal and learning-based controllers (GP, SPGP, LGP, MSGP) for varying mass and inertia.}
\label{fig:tracking_mass_inertia}
\vspace{-0.25cm}
\end{figure*}

Parametric unmodeled dynamics deal with changes or perturbations made to the quadrotor parameters, such as mass or inertia. Parameters are changed to the extent that the nominal controller can still achieve stable flight, although with performance degradation in trajectory tracking. Tracking error is determined in the position space of the quadrotor. The trajectory tracking error of the quadrotor when subjected to changes in the parameters is shown in Figure \ref{fig:tracking_mass_inertia}. In the training phase, the quadrotor is trained for each GP by introducing step changes to the mass and inertia at different time instances, as given below:
\begin{align*}
\hat{m} &= 
\begin{cases} 
      1.00 \cdot m, \qquad \qquad \qquad \qquad \ \ \ \ t_0 \leq t \leq 2 \\
      1.15 \cdot m, \qquad 	\qquad \qquad \qquad \ \ \ \ 2 < t \leq 6 \\
      0.85 \cdot m, \qquad 	\qquad \qquad \qquad \ \ \ \ 6 < t \leq 9 \\
      1.13 \cdot m, \qquad 	\qquad \qquad \qquad \ \ \ \ 9 < t \leq 12 \\
      1.05 \cdot m, \qquad 	\qquad \qquad \qquad \ \ \ \ 12 < t \leq t_f
\end{cases}
\\
\hat{J} &= 
\begin{cases} 
      J + \text{diag}[0.75 \ \ 0.75 \ \ 0.76], \qquad  	t_0 \leq t \leq 2 \\
      J + \text{diag}[0.02 \ \ 0.02 \ \ 0.02], \qquad 		2 < t \leq 6 \\
      J + \text{diag}[1.31 \ \ 1.31 \ \ 1.61], \qquad 		6 < t \leq 9 \\
      J + \text{diag}[0.31 \ \ 0.01 \ \ 0.03], \qquad 		9 < t \leq 12 \\
      J + \text{diag}[0.55 \ \ 0.55 \ \ 0.82], \qquad 		12 < t \leq t_f
\end{cases}
\end{align*}
where $m$ and $J$ are the nominal mass and inertia, $\hat{m}$ and $\hat{J}$ are the perturbed mass and inertia, $t_0 = 0$ and $t_f = 16$ seconds. In the testing phase, the controllers are compared by changing the magnitude of $\hat{m}, \hat{J}$, and time intervals. From the normalized mean squared error (NMSE) plot in Figure \ref{fig:tracking_mass_inertia}, it is clear that the nominal controller has a higher NMSE along $z$. This is expected since changing the mass has more pronounced effect on the altitude. The GP controller performs better than the nominal and SPGP controllers, while LGP outperforms all three controllers. MSGP on the other hand demonstrates superior tracking performance with the lowest NMSE among all the controllers. MSGP achieves better tracking performance compared to the other learning based controllers due to unique hyperparameters and weighted sparse posterior prediction. Moreover, changing dynamical effects at different time instances are better captured with different hyperparameters as opposed to a global set of hyperparameters as in the case of GP, SPGP, and LGP.

\begin{figure*}[!t]
\centering
\includegraphics[width=1\linewidth, height=0.25\linewidth]{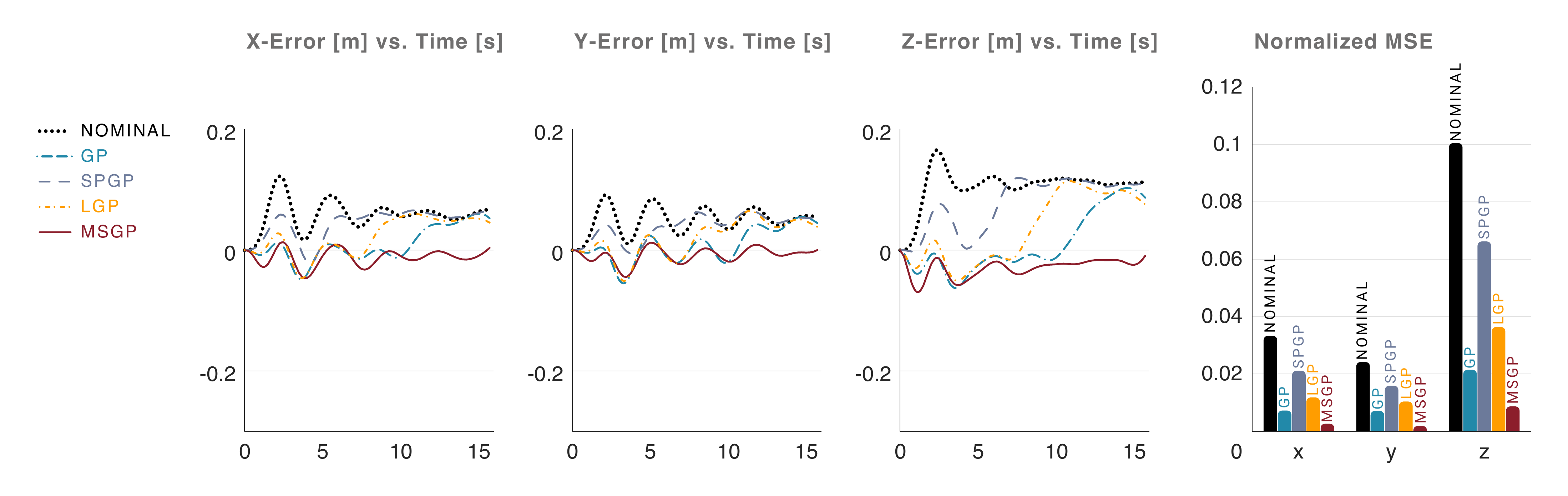}
\vspace{-0.8cm}
\caption{\textit{Non-parametric effects}: Tracking error between nominal and learning-based controllers (GP, SPGP, LGP, MSGP) for varying wind.}
\label{fig:tracking_wind}
\vspace{-0.1cm}
\end{figure*}

\begin{figure*}[!t]
\centering
\includegraphics[width=1\linewidth, height=0.25\linewidth]{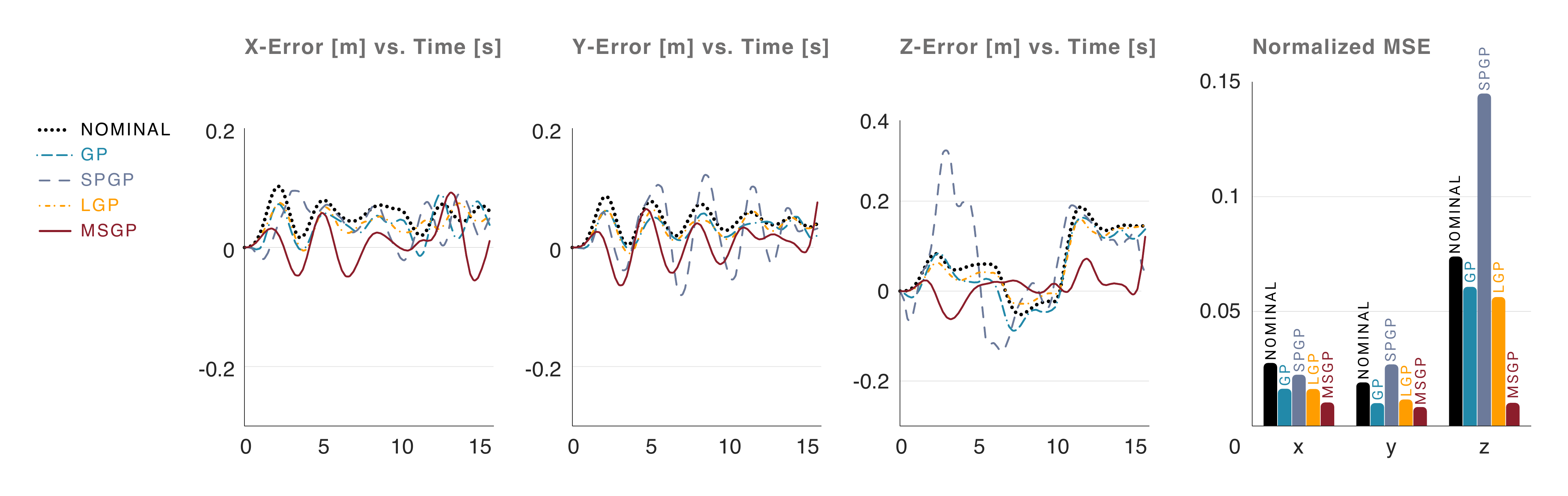}
\vspace{-0.8cm}
\caption{\textit{Parametric \& Non-parametric}: Tracking error between nominal and learning-based controllers (GP, SPGP, LGP, MSGP) for mass, inertia, and wind.}
\vspace{-0.5cm}
\label{fig:tracking_all}
\end{figure*}

\subsection{Non-Parametric Unmodeled Dynamics}\label{subsec:non-param_error}
Here we look at non-parametric effects introduced in the dynamics such as unmodeled aerodynamics. The quadrotor is subjected to the same unknown wind effects for training each GP: $\mathcal{W} = [0.17 \ \ 0.18 \ \ 0.16]^{\top} g$. During testing, a similar wind is introduced having different magnitudes along each dimension for comparing the tracking performance.  Figure \ref{fig:tracking_wind} shows the tracking performance of the quadrotor in presence of non-parametric aerodynamic disturbances.

The nominal controller incurs the highest NMSE along each dimension. This is expected since the unmodeled dynamics cannot be handled by the nominal controller that relies on model knowledge for feedforward compensation. GP performs significantly better than the nominal controller in presence of such effects. SPGP performs better than the nominal case, but it does not compensate as effectively as GP. LGP on the other hand outperforms both the nominal and SPGP controllers, but underperforms compared to GP. Finally, MSGP incurs the lowest NMSE, outperforming all the controllers including GP. 

From the error versus time plots in Figure \ref{fig:tracking_wind}, it can be seen that each learning based controller eventually fails to compensate for the wind effects with the exception of MSGP, which holds out the longest among all the controllers. GP is able to compensate for the wind longer than both SPGP and LGP. SPGP gives in first to the unmodeled dynamical effects since it is only a sparse approximation of GP, while LGP, being a locally clustered approximation of GP, holds out longer than SPGP. Despite MSGP having multiple sparse approximations of GP, it is consistently able to compensate since each sparse model's uniqueness is preserved as described in Section \ref{subsec:hyp_loc}.

\subsection{Parametric \& Non-Parametric Effects}\label{subsec:all_error}
Next, we study the combined effects of unmodeled dynamics in both parametric and non-parametric form. 
Note that the mass, inertia, and wind effects introduced here are different from the previous experiments to show performance against varied conditions. The parametric and non-parametric changes for the training phase are,
\begin{align*}
\hat{m} &= 
\begin{cases} 
      1.00 \cdot m, \qquad \qquad \qquad \qquad t_0 \leq t \leq 6 	\\
      1.23 \cdot m, \qquad \qquad \qquad \qquad 6 < t \leq 10 		\\
      0.81 \cdot m, \qquad \qquad \qquad \qquad 10 < t \leq t_f		
\end{cases}
\\
\hat{J} &= 
\begin{cases} 
	  J + \text{diag}[0.75 \ \ 0.75 \ \ 0.76], \ \ t_0 \leq t \leq 6 \\
      J + \text{diag}[0.52 \ \ 0.52 \ \ 0.52], \ \ 6 < t \leq 10 \\
      J + \text{diag}[1.15 \ \ 1.11 \ \ 1.01], \ \ 10 < t \leq t_f
\end{cases}
\\
\mathcal{W} &= \ [0.31 \ \ 0.32 \ \ 0.15]^{\top} g,	\hspace{1.35cm}    t_0 \leq t \leq t_f
\end{align*}
where $\hat{m}$ and $\hat{J}$ are the affected mass and inertia parameters respectively, $\mathcal{W}$ is the unmodeled wind, $t_0 = 0$ and $t_f = 16$ seconds. Since it is a combination of multiple unmodeled effects, the simulation setup is very challenging because the learning based quadrotor controller needs to deal with a highly inaccurate model. During training, each GP algorithm is trained on the above combinations. During testing, the magnitudes and respective time intervals are altered to test the generalizability of the learning based controllers. 

The tracking error performance is shown in Figure \ref{fig:tracking_all}. Among all the controllers, SPGP performs the worst in terms of tracking error followed by the nominal controller. The sparse approximation tends to over compensate for the introduced nonlinearities in the dynamics, thus exaggerating its effects in the feedforward controller. Both GP and LGP demonstrate comparable performance and perform better than the nominal controller. MSGP has the lowest NMSE among all the controllers with comparable performance to GP and LGP along the $x$ and $y$ dimension. In the altitude domain however, there is a significant reduction in NMSE for MSGP compared to any other controller.

\subsection{Training and Prediction Time Comparison}\label{subsec:sim_time}
We now analyze the average time taken by different GP algorithms for posterior predictions. 
For different training sizes ($3000, 5807, 8613, 11419, 14225$), we train each GP individually subjected to non-parametric wind disturbances. In the case of SPGP, we take one-tenth of each training dataset as pseudo-inputs. For LGP and MSGP, we assume the number of local data points to be linearly proportional to each training dataset. We take one-fifth of each training set to form local models, forming $5$ regional models. Although in practice, the regional models need only have $250-500$ data points each. However, we let each regional model hold a fairly high number of local data points for comparison. Subsequently, for MSGP, we further take one-fifth of each local model's dataset as local pseudo-inputs. For LGP and MSGP, all the neighboring models are considered for computing the weighted posterior prediction, i.e., $N = 5$. 

LGP and MSGP take the least time to train due to the reduced order model compared to GP. LGP takes over $25\si\second$ to train each cluster in CPU time (i7-9800HQ). MSGP on the other hand takes under $6\si\second$ for each cluster in CPU time. We also benchmark GPU training time using the \texttt{GPyTorch} library on a RTX $2080$ Ti \cite{gardner2018gpytorch}. MSGP takes roughly $1 \si\second$ for $12$ dimensional input and $6$ dimensional output for each cluster.

\begin{figure}[!t]
\centering
\includegraphics[width=1\linewidth]{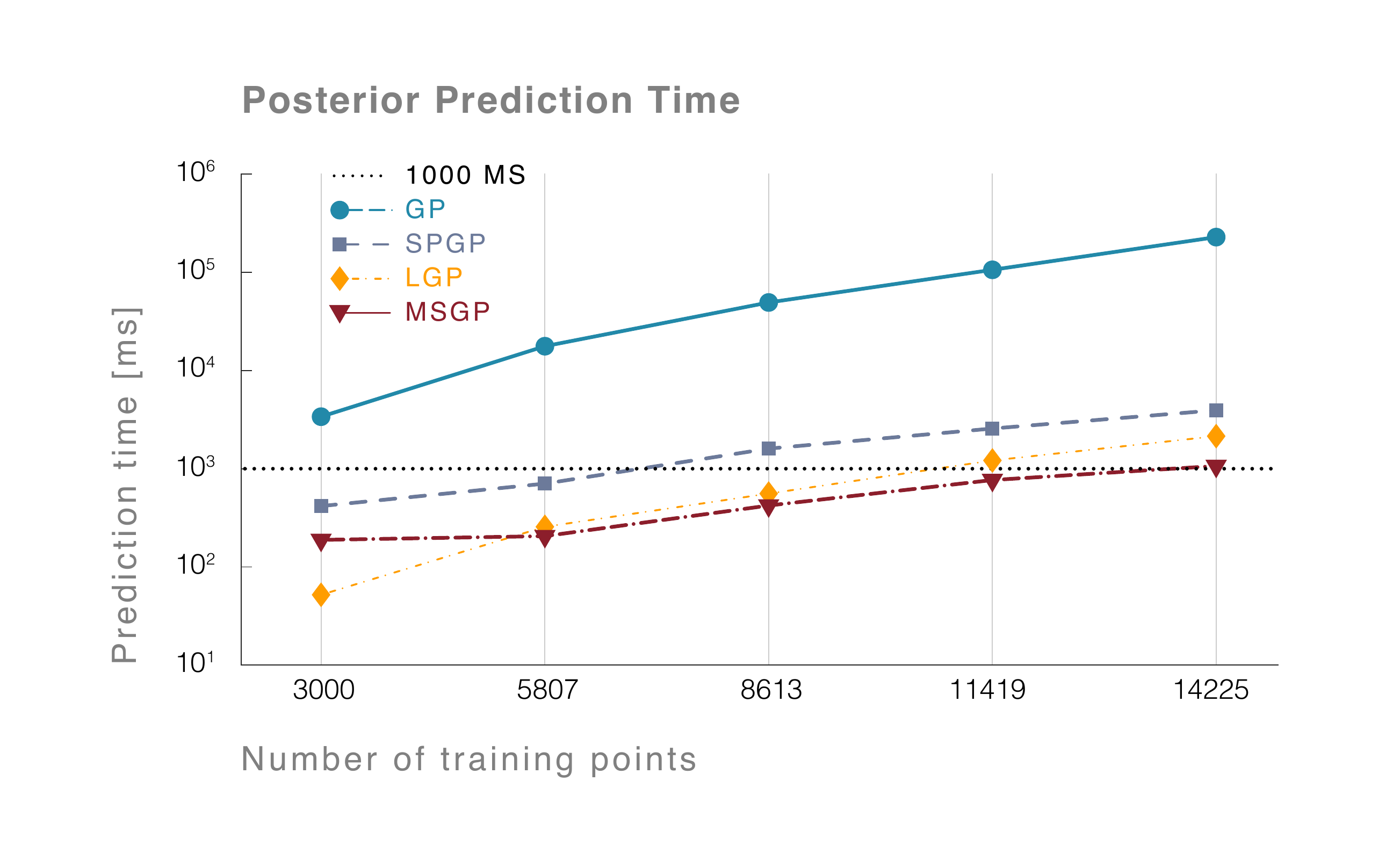}
\vspace{-1cm}
\caption{Posterior prediction time in milliseconds against different training sizes. The prediction time is computed on a query point for GP, SPGP, LGP, and MSGP. The dashed black line marks $1000$ milliseconds.
}
\vspace{-1cm}
\label{fig:time_pred}
\end{figure}

The CPU prediction time comparison is shown in Figure \ref{fig:time_pred}. GP's time complexity drastically increases with increasing training points as expected; since it cubically scales with the number of training points. SPGP scales very well compared to GP. For over $14000$ points, SPGP computes under $5 \si\second$. LGP has the least computational cost with fewer training points (under $5000$) and marginally grows with increasing size. It is faster than SPGP and takes under $2 \si\second$ with over $14000$ points. MSGP is similar to LGP but performs better as the number of training inputs increase due to sparsity in each model. MSGP takes approximately $1 \si\second$  for predictions with over $14000$ training points. Note that precomputations can be made to improve the speed for all the methods. In practice, one can save, $\alpha := (\K+\sigma^2\I)^{-1}y$, where $\K$ denotes the covariance matrix for observed inputs and $\y$ is the set of target observations. Rank-$1$ approximations are then made for computing inverses. This results in tremendous boost in computational speed, thus achieving faster predictions. Doing so results in a prediction time of only $0.8 \si\ms$ for MSGP in CPU time. The space and time complexity for the various GPs are tabulated in Table \ref{tab:complexity}.

\begin{table}[!t]
\begin{tabular}{lcccc} \toprule
{Method} 	& {Storage} 		& {Training} 		& {Mean} 			& {Mean (w/ saving)}	\\ \midrule
\text{GP} 	& $\bigOgpmem$ 		& $\bigOgp$  		& $\bigOgp$ 		& $\bigOgpmean$ 		\\
\text{SPGP}	& $\bigOsgpmem$ 	& $\bigOsgp$  		& $\bigOsgp$ 		& $\bigOsgpmean$		\\
\text{LGP}	& $\bigOlgpmem$ 	& $\bigOgp$  		& $\bigOlgppred$	& $\bigOlgpmean$		\\
\text{MSGP}	& $\bigOmsgpmem$ 	& $\bigOmsgp$  		& $\bigOmsgppred$	& $\bigOmsgpmean$		\\
\bottomrule
\end{tabular}
\caption{Space and time complexity comparison for GP, SPGP, LGP, and MSGP, ignoring the time taken to create $M$ clusters for local methods. The last column assumes saving necessary matrices for each method.
}
\vspace{-0.5cm}
\label{tab:complexity}
\end{table}

\section{HARDWARE EXPERIMENTS}\label{sec:hardware}
\subsection{Experimental Setup}\label{subsec:exp_setup}
We now discuss the implementation of MSGP on a hardware quadrotor platform. We used the Crazyflie 2.1 where state estimation is performed onboard with the help of an external low-cost Lighthouse positioning system \cite{crazyflie}. Control commands are executed from a ground station (Intel i7-7700HQ, 16 GB RAM) through the Crazyradio PA USB dongle. The nominal controller comprises of a position and attitude controller. The position controller generates commanded thrust using a feedforward hovering thrust and feedback PD controller. Desired attitude is maintained through an attitude controller generating commanded roll, pitch, and yaw-rates. The gains selected are, $k_r = 50, k_v = 100$, under which stable flight is maintained within nominal conditions. Control inputs are sent at $100\si\Hz$ while states are recorded at $50\si\Hz$. 
The experiment video can be seen at:

\noindent \url{https://youtu.be/zUk1ISux6ao}.

The objective is to maintain stable flight particularly outside nominal conditions. Training data is collected by adding a  payload of approximately $3 \si\gram$ to the $32 \si\gram$ Crazyflie. MSGP input is $q = [r^{\top}, \dot{r}^{\top}, \ddot{r}^{\top}, \phi, \theta, \psi, F] \in \Rn^{13}$, where $r, \dot{r}, \ddot{r}$, $\phi, \theta, \psi, F$ are positions, velocities, accelerations, roll, pitch, yaw, and commanded thrust. 
Over $5000$ training points were collected, with each sparse model randomly assigned $250$ points. Each model is further sparsed using one-fourth of the points as pseudo-inputs. Training each cluster takes less than $0.16 \si\s$ in CPU time. For weighted sparse prediction, $N = 5$ neighboring models are used.

\subsection{Experiment 1}\label{subsec:exp_altitude}
\begin{figure}[!t]
\centering
\includegraphics[width=1\linewidth]{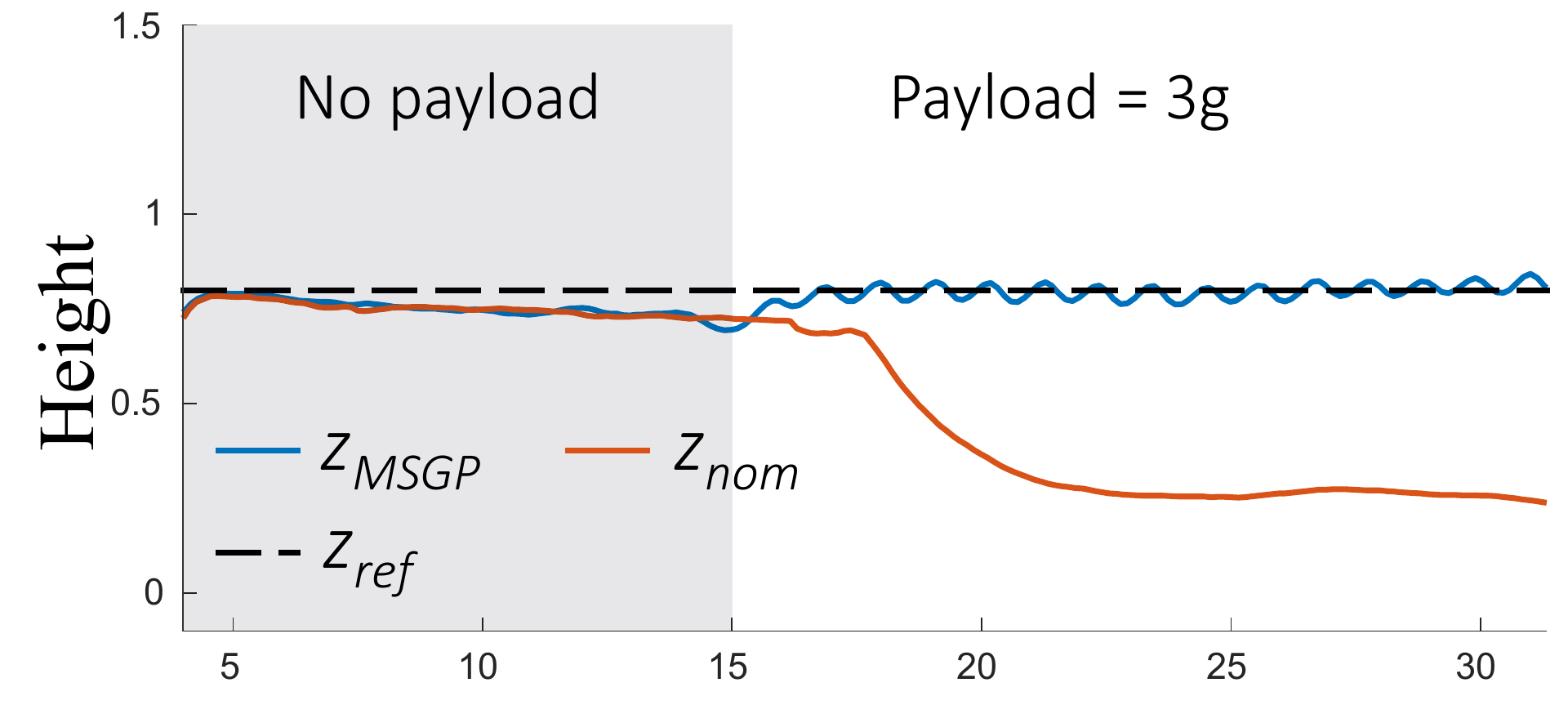}
\vspace{-0.7cm}
\caption{Quadrotor altitude hold with a payload of $3 \si\gram$}
\vspace{-0.25cm}
\label{fig:exp1-altitude}
\end{figure}

\begin{figure}[!t]
\centering
\includegraphics[width=1\linewidth]{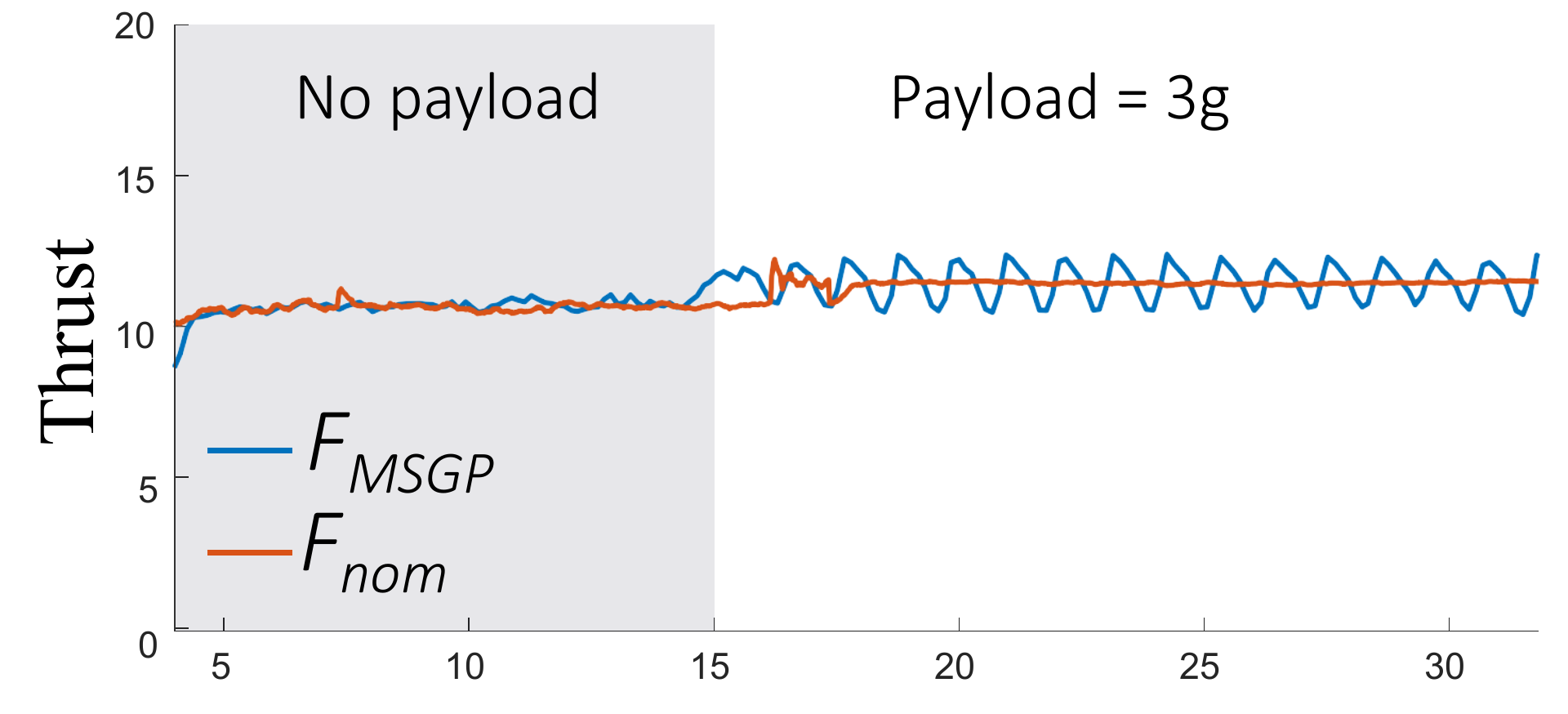}
\vspace{-0.7cm}
\caption{Commanded thrust for altitude hold with a payload of $3 \si\gram$}
\vspace{-0.5cm}
\label{fig:exp1-thrust}
\end{figure}

We first test the nominal and MSGP based controller for a simple task of stable hovering in the presence of an additional payload. The reference altitude is set at $0.8 \si\meter$. Figure \ref{fig:exp1-altitude} shows the tracking comparison for MSGP based controller and nominal controller. Since these are two separate experiments, the transition point for adding the payload is synchronized for visualizing the plot better. When there is no added mass, both MSGP and nominal controller exert similar hovering thrust demonstrating they operate well within nominal conditions. Once the payload is added, MSGP immediately exerts a compensating feedforward thrust as seen in Figure \ref{fig:exp1-thrust}. The nominal controller however begins exerting maximal feedback thrust 
due to the altitude drop experienced. For the nominal controller to eliminate steady state error, the gains need to be adapted or tuned accordingly. This issue is alleviated in the case of the MSGP based controller. We also note that better design of PD gains will reduce the oscillations experienced by MSGP; however, proper design of optimal gains is outside the scope of this work.

\subsection{Experiment 2}\label{subsec:exp_disturbance}
In this experiment, we aggressively disturb the system to test the robustness and generalizability of the MSGP learning algorithm. The system is disturbed in three ways: 1) hitting the payload inducing unmodeled inertial moments, 2) hitting the quadrotor physically off the reference, 3) pulling the quadrotor down with the mass. The disturbances are meant to induce thrust and attitude compensation by MSGP.

\begin{figure}[!t]
\centering
\includegraphics[width=1\linewidth]{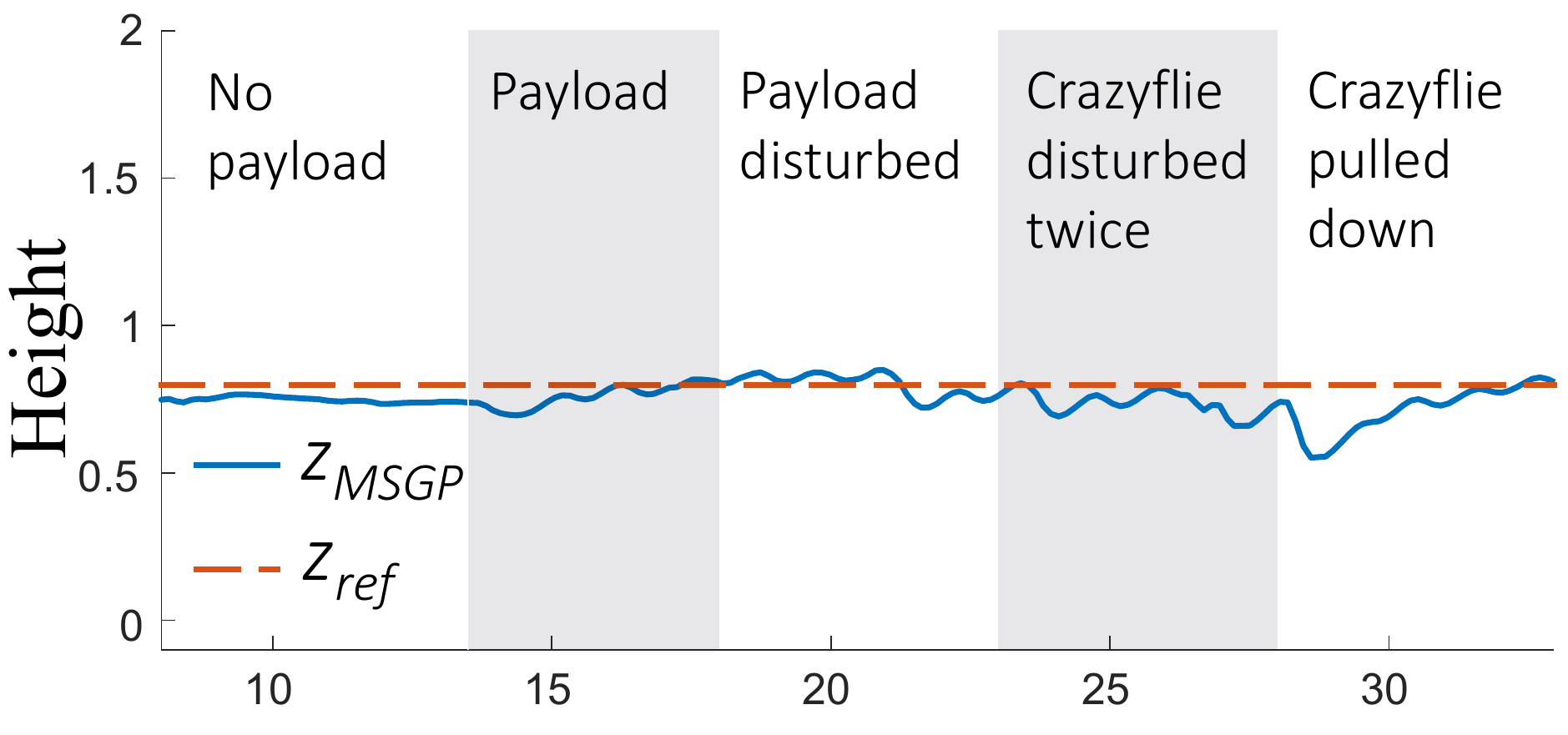}
\vspace{-0.7cm}
\caption{Altitude of the system in presence of aggressive unmodeled disturbances and a payload of $3 \si\gram$ using MSGP controller}
\vspace{-0.4cm}
\label{fig:exp2-altitude}
\end{figure}

\begin{figure}[!t]
\centering
\includegraphics[width=1\linewidth]{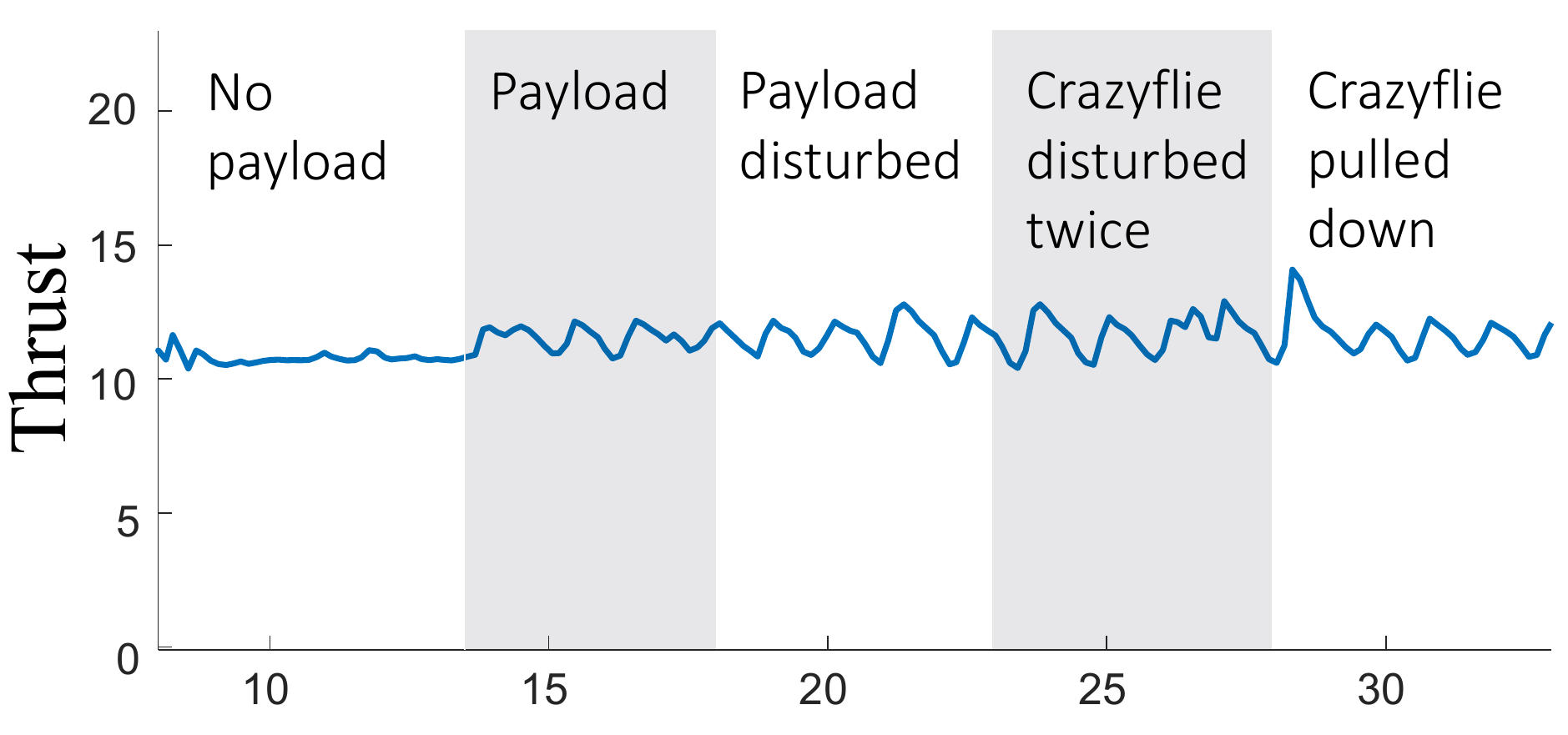}
\vspace{-0.7cm}
\caption{Commanded thrust of the system in presence of aggressive unmodeled disturbances and a payload of $3 \si\gram$ using MSGP controller}
\vspace{-0.5cm}
\label{fig:exp2-thrust}
\end{figure}

The altitude tracking performance in presence of the disturbances is shown in Figure \ref{fig:exp2-altitude}. Note that MSGP has not been trained a priori for these disturbances. MSGP performs good tracking when there is no payload and also compensates well when the mass is added. The transient behavior can be seen in Figure \ref{fig:exp2-thrust} for the feedforward thrust. Thereafter, the system is severely disturbed by first hitting the mass which induces an off-axis inertial moment which MSGP needs to address. Additionally, the Crazyflie is then hit twice to go off its current trajectory. Finally, the Crazyflie is pulled down along with the mass. For all these unmodeled disturbances, MSGP generates the necessary thrust and commanded attitudes to hold stable flight and steady altitude.

\section{CONCLUSION}\label{sec:conclusion}
In summary, we proposed a semi-parametric based control with sparsity by exploiting multiple non-parametric GP sparse models. The sparse models are separately optimized for their hyperparameters, thereby, retaining their own uniqueness. A weighted sparse posterior predictor is adopted for a query point to avoid overfitting and any discontinuities. The proposed framework is tested on a geometric quadrotor controller in simulation with dynamics evolving on the tangent bundle to $SE(3)$. Simulations are performed extensively for unmodeled parametric, non-parametric, and combined dynamical effects for step changes. This demonstrated the proposed approach's efficacy to generalize to step changes despite being a locally sparse approximator. We also rigorously tested our proposed framework against standard GP, sparse GP, and local GP on both prediction quality and time complexity. MSGP demonstrated better prediction accuracy in the form of improved trajectory tracking with reduced prediction time compared to other GPs. Lastly, we experimentally performed sparsity based control on a quadrotor platform. We validated MSGP on the quadrotor dealing with unknown mass and disturbances.



\bibliographystyle{ieeetr}  
\bibliography{root_iros20_msgp}  
\end{document}